\documentclass{article}
\usepackage{ijcai19}

\usepackage{times}
\usepackage{soul}
\usepackage{url}
\usepackage[hidelinks]{hyperref}
\usepackage[utf8]{inputenc}
\usepackage[small]{caption}
\usepackage{graphicx}
\usepackage{amsmath}
\usepackage{multirow}
\usepackage{booktabs}
\usepackage{algorithm}
\usepackage{algorithmic}
\usepackage{latexsym} 
\usepackage{bm}

\title{Adaptive Recurrent Neural Network Based on Mixture Layer}
\author{
Kui Zhao, Yuechuan Li, Chi Zhang, Cheng Yang, Huan Xu
\affiliations
Machine Intelligence Technology Lab, Alibaba Group
\emails
\{zhaokui.zk, yuechuan.lyc, yutou.zc, charis.yangc, huan.xu\}@alibaba-inc.com}

\begin{document}

\maketitle

\begin{abstract}
Although Recurrent Neural Network (RNN) has been a powerful tool for modeling sequential data, its performance is inadequate when processing sequences with multiple patterns. In this paper, we address this challenge by introducing a novel mixture layer and constructing an adaptive RNN. The mixture layer augmented RNN (termed as M-RNN) partitions patterns in training sequences into several clusters and stores the principle patterns as prototype vectors of components in a mixture model. By leveraging the mixture layer, the proposed method can adaptively update states according to the similarities between encoded inputs and prototype vectors, leading to a stronger capacity in assimilating sequences with multiple patterns. Moreover, our approach can be further extended by taking advantage of prior knowledge about data. Experiments on both synthetic and real datasets demonstrate the effectiveness of the proposed method. 
\end{abstract}

\section{Introduction}
Recent years have witnessed great success of deep learning models. 
Owing to the increasing computation resources and strong model capacity, 
neural network models have been applied to numerous applications. 
Among all neural network models, Recurrent Neural Networks (RNNs) \cite{williams1986learning} 
have shown notable potential on sequence modeling tasks, e.g. 
speech recognition \cite{pmlr-v48-amodei16} and 
machine translation \cite{sutskever2014sequence}, 
and therefore receive particular attentions. 
With increasing explorations on RNN, several variants, such as Long Short-Term Memory 
(LSTM) \cite{hochreiter1997long}, 
Gated Recurrent Unit (GRU) \cite{chung2015gated} have been proposed successively. 

The key advantages of RNN come from the recurrent structures,
which carry out the same transition at all time steps, 
and eventually contribute to a satisfactory performance. 
Yet this merit may validate under the assumption that all sequences follow the same pattern.
The conventional RNN may be inappropriate when processing sequences with multiple patterns.
As mentioned in \cite{goodfellow2016deep} \cite{ijcai2017-205} \cite{shazeer2017outrageously}, 
it is difficult to optimize the network when
using the same parameters at all time steps under multiple pattern scenarios.
To this end, more adaptive RNN networks are required.

Recently, some extended mechanisms on RNN are proposed to augment model capability. 
The first one is the attention mechanism \cite{bahdanau2014neural}, 
which is a popular technique in machine translation. 
The attention mechanism suggests aligning data to capture different patterns when translating different words. 
Another attractive mechanism is the Memory-Augmented Neural Networks (MANN) \cite{weston2014memory}, 
whose basic idea is to setup an external memory to memorize the recent and useful samples. 
In this paper, instead of aligning to different parts of current input or memorizing specific samples, 
we introduce a novel mixture layer to memorize principle patterns in training sequences 
and then align the current state to similar patterns in historical data. 

In the proposed mixture layer, patterns in training sequences are partitioned into 
several clusters and their distribution is expressed as a mixture model. 
Consequently, principle patterns can be represented by prototype 
vectors of components in the mixture model, i.e. the centers of clusters. 
When processing input sequences, the mixture layer augmented RNN (termed as M-RNN) 
first measures similarities between the current state and prototype vectors. 
Based on these similarities, the probability that current state belongs to each cluster is 
calculated and we align the current state to historical principle 
patterns by soft assignments. After that, the state in new M-RNN is adaptively updated  
according to the assignments, and therefore it is able to assimilate sequences with multiple patterns. 
In practical situations, 
some prior or domain knowledge about data is often known beforehand, 
and our approach is readily to take advantage of that information to further improve the performance. 
Moreover, although there is a new type of layer added, 
the proposed M-RNN can still be trained by gradient decent in an end-to-end manner. 
To summarize, the contributions of our work are as follows: 
\begin{itemize}
\item We introduce a novel mixture layer, which can be interpreted 
from the mixture model perspective. 
\item By leveraging the mixture layer, we construct M-RNN to adaptively 
process sequences with multiple patterns. 
\item Our approach can be easily extended by combining the prior knowledge about data.
\item Extensive experiments are conducted, including synthetic sequences prediction, 
time series prediction and language modeling. The experimental results 
demonstrate significant advantages of our M-RNN.
\end{itemize}

The remainder of the paper is organized as follows. Section 2 reviews related work.
In Section 3, the mixture layer is introduced in detail, 
as well as its application on LSTM and combination with prior knowledge.   
Experimental evaluations are included in Section 4, 
and final conclusion with looking forward comes in Section 5. 

\section{Related Work}
The research on RNN can be traced back to 1990's \cite{williams1986learning}. 
In the past decades, a number of variants of RNN model appear, in which the most popular one is 
Long Short-Term Memory (LSTM) \cite{hochreiter1997long}. 
LSTM networks introduce memory cells and utilize a gating mechanism to control information flow. 
Another classic RNN variant, Gated Recurrent Unit (GRU), simplifies LSTM with a single update gate, 
which controls the forgetting factor and updating factor 
simultaneously \cite{chung2015gated}. 

Recently, some advanced mechanisms appear to extend RNNs. 
The first one is attention mechanism and particularly useful in machine translation \cite{cho2014learning} \cite{sutskever2014sequence}, 
which requires extra data alignment, and the similarity between the encoded source sentence 
and the output word is calculated. Beyond machine translation, 
the attention mechanism gains notable popularity in other areas including image captioning \cite{chen2018show}, 
rating prediction \cite{cheng20183ncf} and so on. 
A comprehensive study on attention mechanism can be found in \cite{NIPS2017_7181}. 
Instead of aligning to different parts in the current input, our M-RNN aligns 
the current state to similar patterns in historical data.
Moreover, while the attention model is often used in specific scenarios (such as Seq2Seq), 
our mixture layer focuses on solving the problem of modeling multiple patterns and it is a universal block for RNNs. 

Another mechanism is augmented-memory, and its basic idea is to borrow an external memory 
for each sequence \cite{weston2014memory}. 
The  Memory-Augmented Neural Network (MANN) aims to memorize the recent and useful samples, 
and then compare them with the current one in the input sequence. 
The MANN is often used in meta learning and one-shot learning tasks \cite{graves2014neural} \cite{santoro2016meta}, 
and also extended to other applications such as question answering \cite{ijcai2017-280} 
and poetry generation \cite{ijcai2018-633}. 
Different from memorizing specific samples, the mixture layer in our approach 
memorizes principle patterns in training sequences. 

The MOE is a newly introduced layer to extend RNNs by mixture models \cite{shazeer2017outrageously}, 
and it is quite related to our work. In MOE, multiple experts are constructed, 
and the output is controlled by a sparse gating function. 
However, MOE is fairly large and hard to implement without enough resources. 
In contrast, our M-RNN simply introduces a mixture layer to store principle patterns, and it is more concise. 
\section{M-RNN}
\subsection{Mixture Layer}
\label{sc:pm}
\begin{figure}[tbp!]
\centering
\includegraphics[width=6cm]{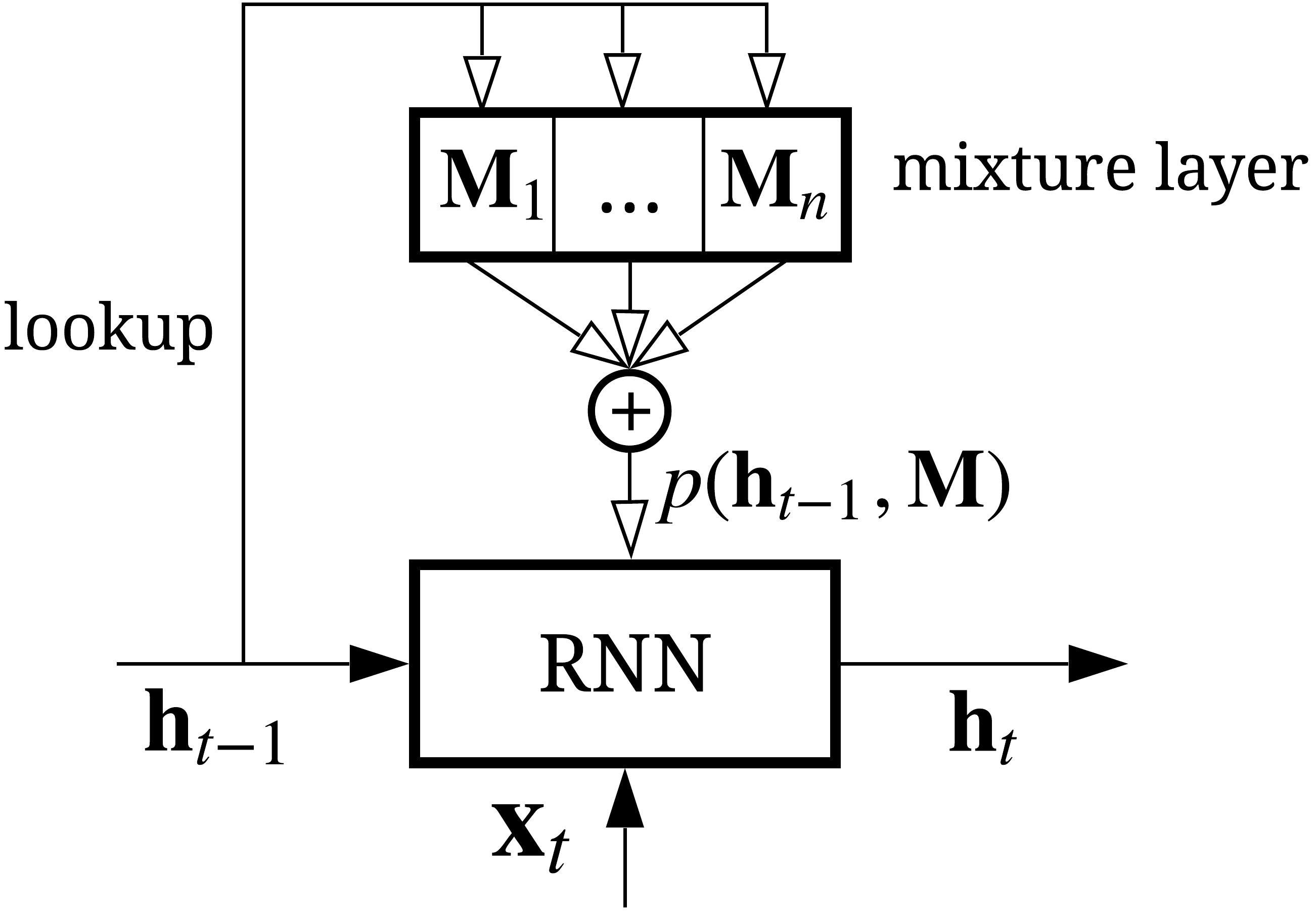}
\caption{The architecture of our MRNN.}
\label{fig:prnn}
\end{figure}

In conventional RNNs, the hidden states are updated with a unique cell at all time steps, which can be expressed as:
\begin{equation}
{\bf h}_{t}=g({\bf h}_{t-1}, {\bf x}_{t}).
\end{equation}
In this paper, a novel mixture layer is introduced and 
a latent matrix ${\bf M}$ (with dimension $m \times n$) is
employed to memorize the principal patterns in training sequences. 
The mixture layer augmented RNN (M-RNN) can flexibly process sequences with multiple patterns.
The structure for M-RNN is illustrated in Figure \ref{fig:prnn}, and the hidden states in M-RNN are formally updated by:
\begin{equation}
{\bf h}_{t}=g({\bf h}_{t-1}, {\bf x}_{t}, p({\bf h}_{t-1}, {\bf M})),
\end{equation}
where $p({\bf h}_{t-1}, {\bf M})$ denotes looking up latent matrix 
and aligning the current state to similar patterns in historical data.
Note that when the mixture layer is introduced, the function $g(\cdot)$ remains similar,
which means the RNN cell need not change its inner structure.
Therefore, our mixture layer can easily equip any RNN cell, and thus it is generally applicable.

The latent matrix ${\bf M}_{m \times n}$ contains $n$ different prototype vectors, 
and each vector represents certain patterns in training sequences.
Given an input sequence $[{\bf x}_1, {\bf x}_2, \dots]$, the hidden state ${\bf h}_{t-1}$ is able to 
represent the subsequence $[{\bf x}_1, \dots, {\bf x}_{t-1}]$. 
And the similarity between ${\bf h}_{t-1}$ and each prototype vector ${\bf M}_i (i \in \{1,\dots, n\})$, 
denoting as $s_i$, can be easily calculated by retrieving each column of the latent matrix. 
This similarity $s_i$ is then used to produce a weight vector ${\bf w}$, with elements computed according to a softmax:  
\begin{equation}
\label{eq:softmax}
w_i = \frac{\exp(s_i)}{\sum\limits_j\exp(s_j)}.
\end{equation}

The weight vector ${\bf w}$ is the strength to amplify or attenuate prototype vectors. 
Consequently, looking up latent matrix and aligning 
the current state to similar historical patterns can be formulated as: 
\begin{equation}
\label{eq:wv}
p({\bf h}_{t-1}, {\bf M})=\sum\limits_i w_i {\bf M}_i.
\end{equation}

In terms of the similarity measure, two alternative approaches are suggested in this section.
The first one is derived from Mahalanobis distance and it is defined as:
\begin{equation}
\label{eq:sed}
s_i =-\frac{1}{2}({\bf h}_{t-1}-{\bf D}{\bf M}_i)^{\rm T}{\bf P}({\bf h}_{t-1}-{\bf D}{\bf M}_i),
\end{equation}
where ${\bf D}$ is the projection matrix from prototype vectors to hidden states, 
and ${\bf P}$ denotes the precision matrix. Another measure is the cosine similarity, 
which is widely applied in related works \cite{graves2014neural} \cite{santoro2016meta} 
considering its robustness and computational efficiency:
\begin{equation}
\label{eq:cos}
s_i = \frac{{\bf h}_{t-1}^{\rm T}{\bf D}{\bf M}_i}{||{\bf h}_{t-1}||_2||{\bf D}{\bf M}_i||_2}. 
\end{equation}

Although there is a new type of layer added, the proposed M-RNN can still be trained 
in an end-to-end manner. That is to say the original parameters in RNN as well as new 
parameters in mixture layer are all updated via gradient decent. 

\subsection{Mixture Model Perspective}
As the latent matrix ${\bf M}$ has $n$ vectors, the mixture layer intends to 
partition all hidden states into $n$ clusters ($z \in \{1,\dots,n\}$). 
Given the hidden state ${\bf h}_{t-1}$, the probability that ${\bf h}_{t-1}$ belongs to the $i$-th cluster is:
\begin{equation}
\begin{aligned}
P(z=i|{\bf h}_{t-1})&=\frac{P(z=i)P({\bf h}_{t-1}|z=i)}{P({\bf h}_{t-1})}\\
&=\frac{P(z=i)P({\bf h}_{t-1}|z=i)}{\sum\limits_j P(z=j)P({\bf h}_{t-1}|z=j)}.
\end{aligned}
\end{equation}

Assume the hidden states ${\bf h}_{t-1}$ are Gaussian variables, then
\begin{equation}
\begin{aligned}
&P({\bf h}_{t-1}|z=i)\\
&=\frac{1}{(2\pi)^{\frac{h}{2}}}\frac{1}{\det(\bm{\Sigma}_i)^{\frac{1}{2}}}
\exp\left(-\frac{1}{2}({\bf h}_{t-1}-{\bm{\mu}}_i)^{\rm T}\bm{\Sigma}_i^{-1}({\bf h}_{t-1}-{\bm{\mu}}_i)\right),
\end{aligned}
\end{equation}
where $h$ is the dimension of ${\bf h}_{t-1}$, ${\bm{\mu}}_i$ and $\bm{\Sigma}_i$ are the 
mean vector and covariance matrix of component $i$ respectively. 

For uniformly distributed prior, i.e. $P(z=i)=\frac{1}{n}$, assume all clusters share the same covariance matrix $\bm{\Sigma}$ ($\bm{\Sigma}^{-1}={\bf P}$) and let ${\bm{\mu}}_i={\bf D}{\bf M}_i$, we have: 
\begin{equation}
\begin{aligned}
&P(z=i|{\bf h}_{t-1})\\
&=\frac{\frac{1}{n}
\exp\left(-\frac{1}{2}({\bf h}_{t-1}-{\bm{\mu}}_i)^{\rm T}\bm{\Sigma}^{-1}({\bf h}_{t-1}-{\bm{\mu}}_i)\right)}
{\sum\limits_j \frac{1}{n}\exp\left(-\frac{1}{2}({\bf h}_{t-1}-{\bm{\mu}}_j)^{\rm T}\bm{\Sigma}^{-1}({\bf h}_{t-1}-{\bm{\mu}}_j)\right)}\\
&=\frac{\exp(s_i)}{\sum\limits_j\exp(s_j)}=w_i,
\end{aligned}
\end{equation}
which is consistent with Eq. (\ref{eq:softmax}) when the similarity measure follows Eq. (\ref{eq:sed}). 
Thus, aligning the current state to similar historical patterns 
is equal to the soft assignment of ${\bf h}_{t-1}$ to specific components in the mixture model: 
\begin{equation}
\begin{aligned}
p'({\bf h}_{t-1}, {\bf M})&=\sum\limits_i P(z=i|{\bf h}_{t-1}){\bm{\mu}}_i\\
&=\sum\limits_i w_i {\bf DM}_i={\bf D}p({\bf h}_{t-1}, {\bf M}).
\end{aligned}
\end{equation}
Let ${\bf D}^+$ be the Moore-Penrose inverse of ${\bf D}$, we have:
\begin{equation}
p({\bf h}_{t-1}, {\bf M})={\bf D}^+p'({\bf h}_{t-1}, {\bf M})=\sum\limits_i w_i {\bf M}_i,
\end{equation}
and thus looking up latent matrix in Eq. (\ref{eq:wv}) can be interpreted from the mixture model perspective.

As a matter of fact, looking up and updating the latent matrix can be seen as a variant of EM algorithm. 
For given hidden state ${\bf h}_{t-1}$ and current latent matrix ${\bf M}$, E-step assigns ``responsibility'' $w_i$ 
to each cluster via measuring similarities in forward propagation. 
The M-step optimizes the parameters in prototype vectors based on $w_i$ in backward propagation. 
Rather than explicit parameter updating mechanism, which appears in conventional EM procedure 
for Gaussian mixture model, the update of mixture layer is implicit and based on gradient descent. 

When cosine similarity is selected instead of the similarity in Eq. (\ref{eq:sed}), 
and ${\bf h}_{t-1}$ follows Von Mises-Fisher distribution ($\kappa=1$) 
with the following probability density function:
\begin{equation}
\begin{aligned}
\label {eq:vmf}
P({\bf h}_{t-1}|z=i)=C\exp\left({\bf h}_{t-1}^{\rm T}{\bm{\mu}}_i\right),
\end{aligned}
\end{equation}
where ${\bm{\mu}}_i = {\bf DM}_i$ and $C$ is the normalization constant, we also have $P(z=i|{\bf h}_{t-1})=w_i$, 
and thus it can be interpreted from the mixture model perspective as well. 

\subsection{LSTM with Mixture Layer}
As a general mechanism, the mixture layer is able to equip almost all RNN models. 
In this section, we take LSTM as an example and illustrate how the mixture layer works in practice.
Since the retrieved result of looking up latent matrix in mixture layer is added to all gates and cells, 
the forget gate ${\bf f}_t$ and the input gate ${\bf i}_t$ are revised as: 
\begin{equation}
{\bf f}_t=\sigma({\bf W}_f[{\bf h}_{t-1}, {\bf x}_t, p({\bf h}_{t-1}, {\bf M})] + {\bf b}_f).
\end{equation}
\begin{equation}
{\bf i}_t=\sigma({\bf W}_i[{\bf h}_{t-1}, {\bf x}_t, p({\bf h}_{t-1}, {\bf M})] + {\bf b}_i).
\end{equation}
Then the memory cell ${\bf c}_t$ can be updated adaptively: 
\begin{equation}
\tilde{\bf c}_t=\tanh({\bf W}_c[{\bf h}_{t-1}, {\bf x}_t, p({\bf h}_{t-1}, {\bf M})] + {\bf b}_c).
\end{equation}
\begin{equation}
{\bf c}_t={\bf f}_t\circ{\bf c}_{t-1} + {\bf i_t}\circ \tilde{\bf c}_t.
\end{equation}
Consequently, the output gate ${\bf o}_t$ and cell ${\bf h}_t$ become: 
\begin{equation}
{\bf o}_t=\sigma({\bf W}_o[{\bf h}_{t-1}, {\bf x}_t, p({\bf h}_{t-1}, {\bf M})] + {\bf b}_o).
\end{equation}
\begin{equation}
{\bf h}_t = {\bf o}_t \circ \tanh({\bf c}_t).
\end{equation}

\subsection{Mixture Layer with Prior Knowledge}
In many practical situations, some prior or domain knowledge, 
about sample distribution or data pattern, is known beforehand.
As a matter of fact, the prior knowledge can provide much information
and help to build more efficient mixture layer, which substantially benefits the training process.
In this subsection, we select an example in text modeling for illustration. 
The category information (e.g. sports, science) is often provided in advance in text modeling. 
To leverage this prior knowledge, we set up multiple latent matrices for RNN.
Assume there exist $B$ different buckets (categories) in the training data, 
a unique latent matrix ${\bf M}^{k}$ ($k \in \{1, \dots B\}$) will be allocated for each bucket.
Therefore, Eq. (\ref{eq:wv}) is extended to be:
\begin{equation}
p({\bf h}_{t-1}, {\bf M}^{k}) = \sum\limits_i w_i {\bf M}_i^{k}.
\end{equation}
The samples are partitioned according to bucket information, 
and specific samples are utilized to update each latent matrix.

\section{Experiments}
The performance of our M-RNN is evaluated on three different tasks: synthetic sequences prediction, 
time series prediction and language modeling. 
Experiments are implemented in Tensorflow \cite{abadi2016tensorflow}. 
As two different measures for calculating similarities are suggested in above sections 
and their performances are very close in practice, 
we only report results under cosine similarities for simplicity. 
It is claimed that for any baseline model (e.g. LSTM), 
we name the mixture layer augmented model with prefix M (e.g. M-LSTM), 
and name the prior knowledge based mixture layer augmented model with prefix PM (e.g. PM-LSTM).

All trainable parameters are initialized randomly from a uniform distribution over [-0.05, 0.05]. 
The parameters are updated through back propagation with 
Adam rule \cite{kingma2014adam} and the learning rate is 0.001.
When tuning the hyper-parameters (e.g. the dimension $m$ of vector and the number $n$ of clusters in mixture model), 
an independent validation set is randomly drawn from the training set, 
and the model is trained on the remaining samples. After the hyper-parameters have been determined, 
the model is trained again on entire training set. 
All experiments are run several times and the average results are reported. 

\subsection{Synthetic Sequences Prediction}
\subsubsection{Dataset.}
\begin{figure}[tbp]
\begin{center}
\begin{minipage}[t]{0.99\linewidth}
\centering
\includegraphics[width=3.in]{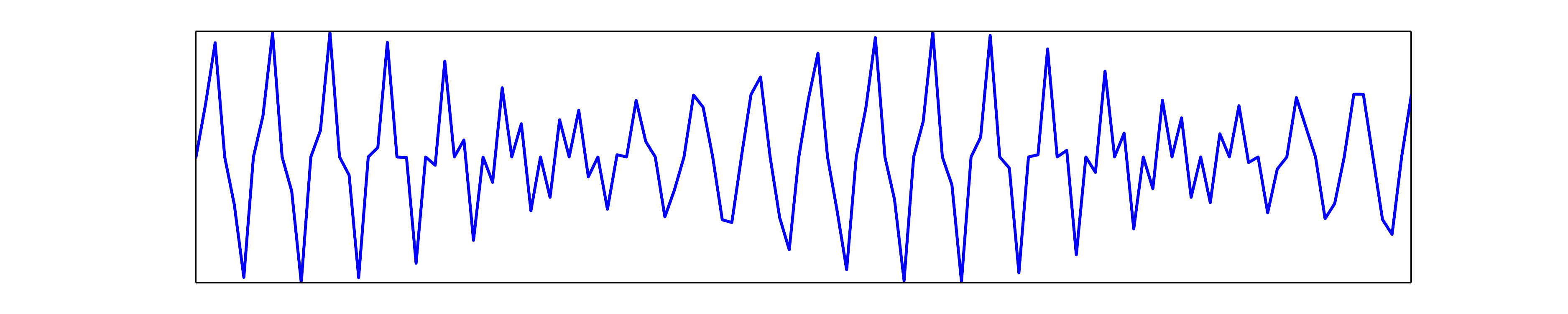}
\end{minipage}\\
\begin{minipage}[t]{0.99\linewidth}
\centering
\includegraphics[width=3.in]{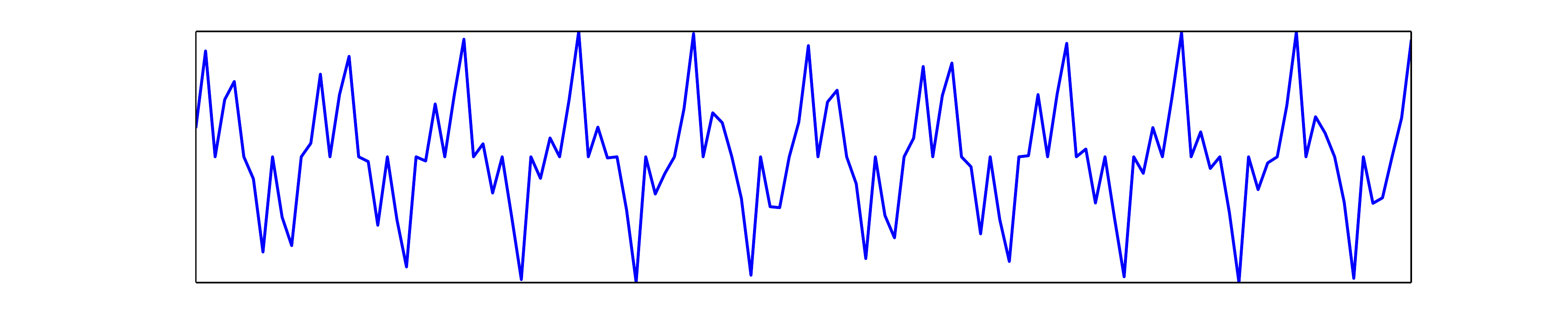}
\end{minipage}
\begin{minipage}[t]{0.99\linewidth}
\centering
\includegraphics[width=3.in]{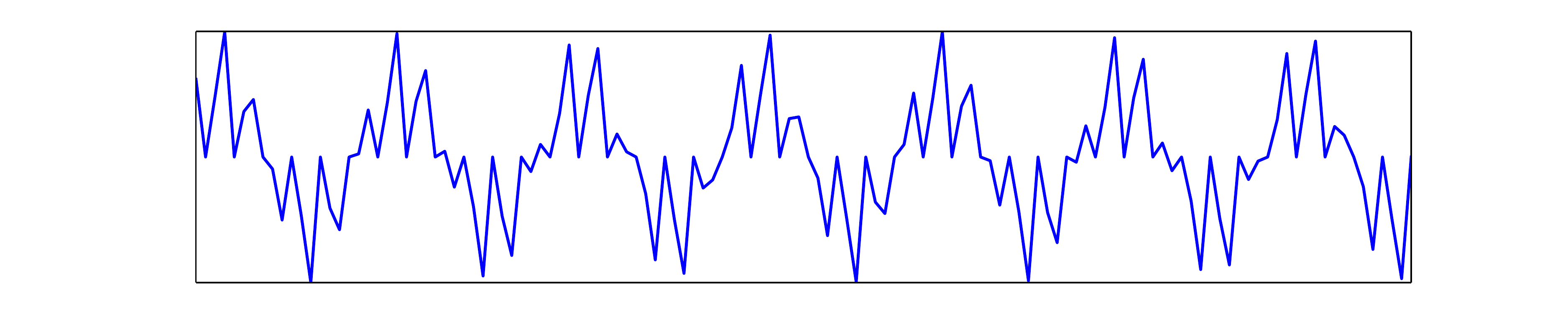}
\end{minipage}
\caption{Three samples of synthetic sequences. }
\label{fig:seq}
\end{center}
\end{figure}

We conduct sequences prediction on a synthetic dataset to show how mixture layer helps 
to learn from data with multiple patterns. 
Let $N$ be the number of generated sequences and $M$ be the length of each sequence. 
The $j$-th $(j=1,\dots, M)$ elements in $i$-th $(i=1,\dots, N)$ sequence is calculated as:
\begin{equation}
\label{eq:ssp}
s_{ij}=(i+j)\text{ mod } 3 \times \sin\left(\frac{i+j}{i\text{ mod } 3 + 1}\right), 
\end{equation}
where $\mod$ denotes the modulo operator, 
and some samples are shown in Figure \ref{fig:seq}. 
For each sequence $s_i$, the task is to predict the last element $s_{iM}$ when $[s_{i1},\dots, s_{iM-1}]$ is given. 
In the experiment, $N$ is set to be 25600 and $M$ is set to be $128$, and one half of sequences are randomly 
selected as the testing set. 

\subsubsection{Setup.}
We select LSTM as baseline and add the mixture layer to construct our M-LSTM. 
From Eq. (\ref{eq:ssp}), we can see that there are three different types of cycles, 
and we accordingly construct the prior knowledge based M-LSTM (termed as PM-LSTM). 
The LSTM has a single layer where the number of hidden units equals to 
8. The dimension of latent matrix in M-LSTM is $4\times 3$, and a unique ${\bf M}_{4\times 3}$ is assigned 
to each type of cycle in PM-LSTM. The models are trained for 10 epochs. 

\subsubsection{Results.}

\begin{table}[!tbp]
\caption{The MAE results on synthetic sequences prediction tasks (lower is better).}
\begin{center}
\begin{tabular}{c|c}
\toprule
Model & MAE\\
\hline
LSTM & 0.090 \\
M-LSTM & 0.076\\
PM-LSTM & {\bf 0.026} \\
\bottomrule
\end{tabular}
\end{center}
\label{tb:ss}
\end{table}
The results are measured by Mean Absolute Error (MAE), and it is written as:  
\begin{equation}
\text{MAE}=\frac{\sum_{i=1}^{N}|y_i-\hat{y}_i|}{N},
\end{equation}
where $N$ is the number of testing samples, $y_i$ is the true value and $\hat{y}_i$ is the prediction result. 
All MAE results are shown in Table \ref{tb:ss}. 
We can see that mixture layer 
can significantly enhance the capability of LSTM in processing sequences with multiple patterns. 
Moreover, it is readily to leverage prior knowledge to further enhance the performance. 

\subsection{Time Series Prediction}
\subsubsection{Dataset.}
We conduct time series prediction experiments on two datasets: Power Consumption (PC) and Sales Forecast (SF). 
\begin{itemize}
\item {\bf Power Consumption (PC)} \cite{Lichman:2013}. 
This dataset contains measurements of electric power consumption in one household over a period 
of 198 weeks\footnote{\url{https://archive.ics.uci.edu/ml/datasets/Individual+household+electric+power+consumption}}. 
The global-active-power is aggregated into hourly averaged time series, 
and the prediction target is the global-active-power at every hour on the next day. 
Each day is divided into two periods: high consumption time (7:00-13:00 and 18:00-22:00) 
and low consumption time (otherwise), which can be utilized as the prior knowledge.
The entire dataset is separated into two parts: 
training part (dates range [2007-04-08, 2010-11-19]) 
and testing part (dates range [2010-11-20, 2010-11-26]). 

\item {\bf Sales Forecast (SF). }
This dataset is collected from one of the largest E-commerce platforms in the world, 
and it contains four features (i.e. browse times, customer number, price and sales) of
1 million items from 769 categories over 13 weeks.
The target is to predict the total sales in the next week for each item. 
As the category information is provided, it can be utilized as the prior knowledge.
The training set includes over 3 million samples,
and the size of testing set is about 2 thousand (testing sample list is provided by large merchants).
\end{itemize}

\subsubsection{Setup.} 
We select LSTM as baseline model. 
The LSTM is first compared with the classical ARIMA model \cite{hamilton1994time}. 
Then the mixture layer is added to construct our approaches. 
The hyper-parameters are set up on two datasets as follows respectively: 

\begin{itemize}
\item {\bf Power Consumption (PC).} 
For each hour, the input sequence contains the related power consumption on the last 56 days. 
At every step, the sequence contains values at three adjacent hours centering at the target hour. 
The goal is to predict global-active-power at every hour on the next day. 
The LSTM has a single layer where the dimension of hidden units is 32. 
In terms of our approach, the M-LSTM is constructed 
by adding a mixture layer with ${\bf M}_{4 \times 8}$.
To construct the prior knowledge based M-LSTM (termed as PM-LSTM), 
a unique ${\bf M}_{4 \times 8}$ is assigned to each 
period (high consumption or low consumption).  
The models are trained for 30 epochs.

\item {\bf Sales Forecast (SF).}  For each item, the input is a sequence of four features on recent 56 days, 
and the target is to predict total sales in the next week. 
The LSTM has a single layer where the dimension of hidden units is 64. 
Similarly, our M-LSTM is constructed by adding a mixture layer with ${\bf M}_{8 \times 16}$.
By taking the category information as prior knowledge, 
we assign a unique ${\bf M}_{8 \times 16}$ to each category and construct the PM-LSTM model.
The models are trained for 15 epochs. 
\end{itemize}  

\subsubsection{Results.}
\begin{table}[!tbp]
\caption{The RMAE results on time series prediction tasks (lower is better).}
\begin{center}
\begin{tabular}{c|c|c}
\toprule
& Model & RMAE (\%) \\
\hline
\multirow{3}{*}{PC}
& ARIMA & 40.2 \\
& LSTM & 35.4 \\
& M-LSTM & 34.4 \\
& PM-LSTM & {\bf 33.9} \\
\hline
\hline
\multirow{4}{*}{SF}
& ARIMA & 84.2 \\
& LSTM & 56.8  \\
& M-LSTM & 52.5 \\
& PM-LSTM & {\bf 42.9} \\
\bottomrule
\end{tabular}
\end{center}
\label{tb:ts}
\end{table}
The results are measured in Relative Mean Absolute Error (RMAE), and it is written as:  
\begin{equation}
\text{RMAE}=\frac{\sum_{i=1}^{N}|y_i-\hat{y}_i|}{\sum_{i=1}^{N}y_i},
\end{equation}
where $N$ is the number of testing samples, $y_i$ is the true value and $\hat{y}_i$ is the prediction result. 
All RMAE results are shown in Table \ref{tb:ts}. 
It is evident to see that multiple patterns commonly exist in real-world data. 
Thus our M-LSTM model significantly outperforms LSTM and ARIMA, on both PC and SF datasets. 
Therefore, the advantages of our mixture layer is fully demonstrated.
In addition, by leveraging prior knowledge,
the PM-LSTM is able to further enhance the prediction accuracy on both PC and SF datasets,
which indicates the strong adaptability of our approach as well.

\subsection{Language Modeling}
\subsubsection{Dataset.}
We conduct word-level prediction experiments on 20NG: 

\begin{itemize}
\item {\bf 20-Newsgroup (20NG)} \cite{lang1995newsweeder}. 
This dataset is originally a benchmark for text categorization, 
where 20 thousand news documents are evenly categorized into 20 groups. 
In word-level prediction, the group information is considered as prior knowledge. 
The preprocessed data can be found in \cite{2007:phd-Ana-Cardoso-Cachopo}\footnote{\url{http://ana.cachopo.org/datasets-for-single-label-text-categorization}},
and it consists of 5 million words and the vocabulary contains 74 thousand words. 
The most frequent 10 thousand words are selected as the final vocabulary in our experiment. 
\end{itemize}

\subsubsection{Setup.}
There are two different settings in the experiments: simple setting and complex setting.
As the purpose of simple setting is to demonstrate the benefit of mixture layer,  
the complex setting extends the advantages of our approach to more complicated models. 

\begin{itemize}
\item {\bf Simple setting.} In this setting, each word is first embedded into a 32-dimensional vector. 
The size of hidden units in LSTM is 128. 
The dimension of latent matrix in M-LSTM is 
${16\times 10}$. The group information is utilized as prior knowledge, 
and we construct PM-LSTM by allocating a unique ${\bf M}_{16\times 10}$ for each group. 
The model is trained for 20 epochs. 

\item {\bf Complex setting.} The ``large'' network architecture in \cite{zaremba2014recurrent} 
is one of the state-of-art models and provides a strong baseline for language modeling 
tasks\footnote{An open source implementation: 
\url{https://github.com/tensorflow/models/blob/master/tutorials/rnn/ptb/ptb_word_lm.py}}. 
This model contains many extensions, including multiple layers stacking, dropout, gradient clipping, 
learning rate decay and so on. In this setting, we select the ``large'' network (named by LARGE) as the baseline model. 
In terms of our approach, a mixture layer with ${\bf M}_{512\times 32}$ 
is added into LARGE to construct our M-LARGE model. 
Similarly, the prior knowledge based M-LARGE (termed as PM-LARGE) is established by 
introducing a particular ${\bf M}_{512\times 32}$ for each group.
\end{itemize}

With more sophisticated model \cite{zilly2016recurrent} or ensemble of multiple models, 
lower perplexities can be achieved. However, we here simply focus on assessing the impact of 
mixture layer when added to existing architectures, rather than the most absolute state-of-the-art performance. 

\subsubsection{Results.}
\begin{table}[!tbp]
\caption{The word-level perplexity results on language modeling tasks (lower is better).}
\begin{center}
\begin{tabular}{c|c||c|c}
\toprule
Simple & Perp. & Complex & Perp.\\
\hline
LSTM & 178.9& LARGE & 109.1  \\
M-LSTM & 168.2 & M-LARGE & 108.4 \\
PM-LSTM & {\bf 157.1}& PM-LARGE & {\bf 105.4} \\
\bottomrule
\end{tabular}
\end{center}
\label{tb:lm}
\end{table}
All results are measured in perplexity, which is a popular metric to evaluate language 
models \cite{katz1987estimation}, and the results are summarized in Table \ref{tb:lm}.  
In the simple setting, the benefit from the mixture layer is significant.
The content in 20NG is semantically rich, and our model can easily distinguish various semantic patterns in the dataset. 
Similarly, with the assist from prior knowledge, the efficiency of mixture layer is further improved.
The advantages of our approach remain in the complex setting. 
We can see that the LARGE model has already achieved a high accuracy, 
and our approach is able to further enhance the model performance.
Thus, the general superiority of our approach is proved under both simple setting and complex setting.

\subsection{Discussion}
\subsubsection{Other mechanisms analysis.}
We also conduct some experiments on other mechanisms, including GRU 
and attention mechanism, and all results are are summarized in Table \ref{tb:dis}. 
GRU is another well-known RNN variant and we test the augmented 
GRUs (termed as M-GRU and PM-GRU) on 20NG dataset under the simple setting. 
As we can see, although GRU has better performance on this task, 
the mixture layer still significantly enhances its performance, 
and thus the proposed mixture layer is a universal block for RNNs.
The attention mechanism is often used in specific scenarios, such as Seq2Seq, 
and we have tried to apply it in language modeling experiments. 
However, the attention method cannot achieve satisfying results, 
and the testing perplexity on 20NG dataset is only 558.6, 
which shows the attention mechanism may be not suitable for such tasks. 

\subsubsection{Model size analysis.}
After introducing the mixture layer, the number of parameters increases, 
and the comparison seems unfair at first glance. 
However, the increase of model size caused by mixture layer is very small. 
Taking the simple setting in language model for illustration, 
the original LSTM has 82432 parameters in the cell, 
and 1362432 parameters in the entire network (cell parameters plus parameters in 10000-classes sigmoid function). 
After introducing mixture layer, the number of newly added parameters is 10400. 
So the increment in model size is 0.76\%, while the word-level perplexity decreases 6\%. 
Therefore, the improvement is significant and the introduction of mixture layer is economic.
To better understand the effect of increasing model size, we conduct an additional experiment on 20NG dataset. 
By increasing the number of hidden units, we make the number of parameters in cell of new LSTM almost equal 
to that in cell of M-LSTM. This new LSTM has much more parameters than our M-LSTM 
due to larger linear projection, and the comparison is quite unfair for our M-LSTM. 
However, the perplexity achieved by new LSTM is 176.9, which is still far away from our M-LSTM (168.2). 
Obviously, simply increasing the model size cannot effectively enhance the performance.

\subsubsection{Mixture model convergence.}
\begin{table}[!tbp]
\caption{The word-level perplexity results of other mechanisms on 20NG dataset under simple setting (lower is better).}
\begin{center}
\begin{tabular}{c|c}
\toprule
 Model & Perplexity\\
\hline
GRU & 174.3\\
M-GRU & 163.7\\
PM-GRU & 152.9\\
Attention & 558.6 \\
\bottomrule
\end{tabular}
\end{center}
\label{tb:dis}
\end{table}

\begin{figure}[tbp!]
\centering
\includegraphics[width=7.5cm]{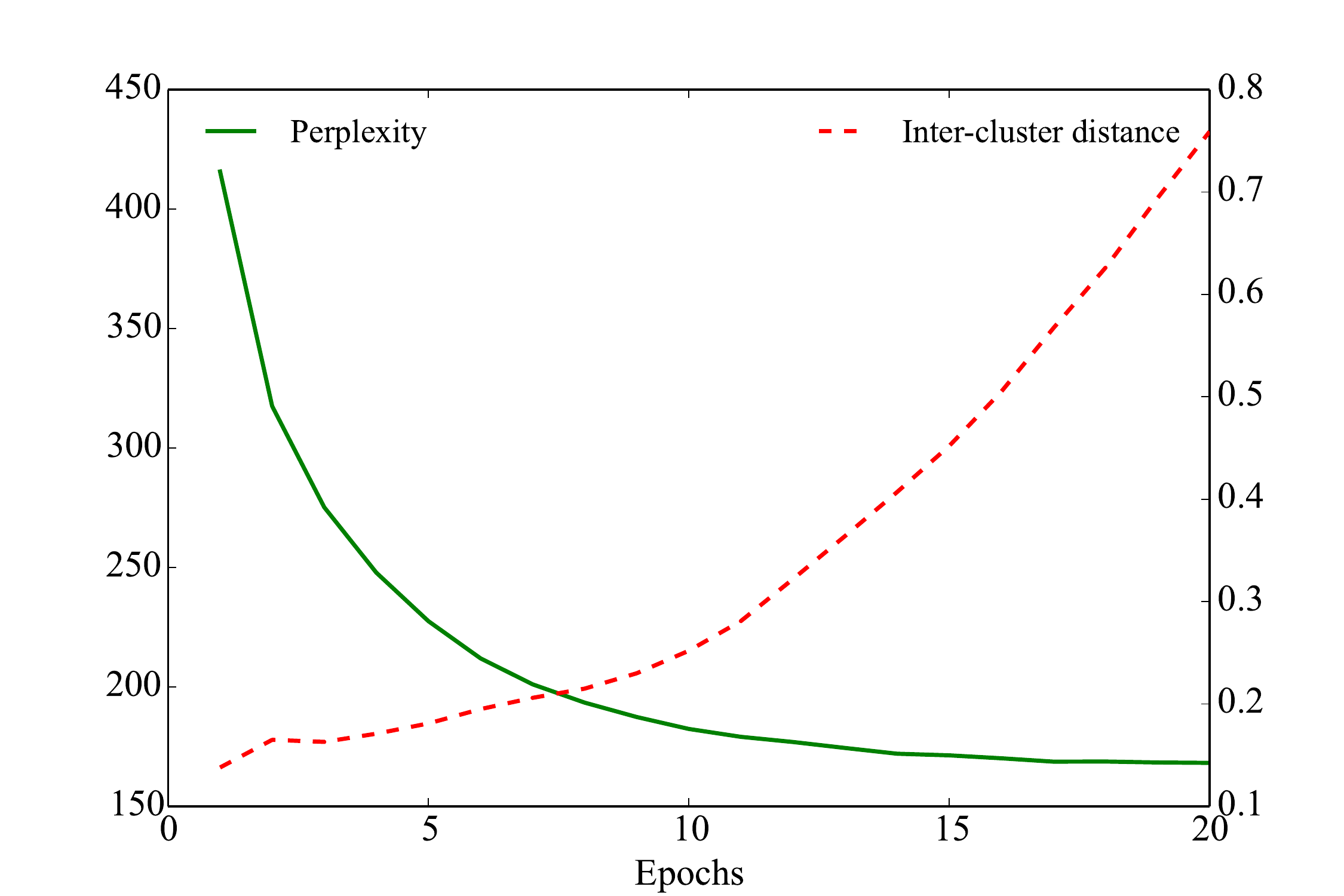}
\caption{The changing process of average Euclidean distance among all centers.}
\label{fig:edc}
\end{figure}
As mentioned before, the mixture layer can be interpreted from mixture model perspective, 
and the mixture layer stores each principle pattern as the center of each cluster (or prototype vector). 
To gain more insights about latent matrix, we take the simple setting on 20NG dataset 
as an example, and plot the change of average Euclidean distance 
among all centers during the training procedure in Figure \ref{fig:edc}. 
The green solid line denotes the word-level perplexity on testing set 
and the red dotted line is the average Euclidean distance. 
As we can see, all vectors are similarly initialized at the beginning of training, 
and they gradually become isolated as the model converges, 
which means each cluster center represents a distinct pattern in the training data.

\section{Conclusion}
Conventional RNN is limited to adaptively process sequences with multiple patterns.
In this paper, an adaptive RNN is proposed and a novel mixture layer is 
introduced to memorize the principal patterns in training sequences by a mixture model.
By aligning the current state to similar patterns in historical data, 
the mixture layer augmented RNN (termed as M-RNN) applies adaptive transition at each time step. 
Moreover, the proposed approach can easily utilize the prior knowledge about data. 
Although there is a new type of layer added, 
the entire network can still be trained by gradient decent in an end-to-end manner. 
Experiments on extensive tasks demonstrate the effectiveness and superiority of our approach.

The proposed mixture layer is a universal block, and we look forward to applying it to other types of neural networks, 
such as feed-forward networks and convolutional neural networks. 
Another interesting but challenging topic for further studies is adding sparsity 
to the weight vector computed by Eq. (\ref{eq:softmax}), which is helpful to boost the computation. 

\bibliographystyle{named}
\bibliography{ijcai19}

\begin{thebibliography}{}

\bibitem[\protect\citeauthoryear{Abadi \bgroup \em et al.\egroup
  }{2016}]{abadi2016tensorflow}
Mart{\'\i}n Abadi, Ashish Agarwal, Paul Barham, Eugene Brevdo, Zhifeng Chen,
  Craig Citro, et~al.
\newblock Tensorflow: Large-scale machine learning on heterogeneous distributed
  systems.
\newblock {\em arXiv preprint arXiv:1603.04467}, 2016.

\bibitem[\protect\citeauthoryear{Bahdanau \bgroup \em et al.\egroup
  }{2014}]{bahdanau2014neural}
Dzmitry Bahdanau, Kyunghyun Cho, and Yoshua Bengio.
\newblock Neural machine translation by jointly learning to align and
  translate.
\newblock {\em arXiv preprint arXiv:1409.0473}, 2014.

\bibitem[\protect\citeauthoryear{Cao \bgroup \em et al.\egroup
  }{2017}]{ijcai2017-205}
Zhu Cao, Linlin Wang, and Gerard de~Melo.
\newblock Multiple-weight recurrent neural networks.
\newblock In {\em Proceedings of the International Joint Conference on
  Artificial Intelligence}, pages 1483--1489, 2017.

\bibitem[\protect\citeauthoryear{Cardoso-Cachopo}{2007}]{2007:phd-Ana-Cardoso-Cachopo}
Ana Cardoso-Cachopo.
\newblock Improving methods for single-label text categorization.
\newblock PdD Thesis, Instituto Superior Tecnico, Universidade Tecnica de
  Lisboa, 2007.

\bibitem[\protect\citeauthoryear{Chen \bgroup \em et al.\egroup
  }{2018}]{chen2018show}
Hui Chen, Guiguang Ding, Zijia Lin, Sicheng Zhao, and Jungong Han.
\newblock Show, observe and tell: Attribute-driven attention model for image
  captioning.
\newblock In {\em Proceedings of the International Joint Conference on
  Artificial Intelligence}, pages 606--612, 2018.

\bibitem[\protect\citeauthoryear{Cheng \bgroup \em et al.\egroup
  }{2018}]{cheng20183ncf}
Zhiyong Cheng, Ying Ding, Xiangnan He, Lei Zhu, Xuemeng Song, and Mohan~S
  Kankanhalli.
\newblock A\^{} 3ncf: An adaptive aspect attention model for rating prediction.
\newblock In {\em Proceedings of the International Joint Conference on
  Artificial Intelligence}, pages 3748--3754, 2018.

\bibitem[\protect\citeauthoryear{Cho \bgroup \em et al.\egroup
  }{2014}]{cho2014learning}
Kyunghyun Cho, Bart Van~Merri{\"e}nboer, Caglar Gulcehre, Dzmitry Bahdanau,
  Fethi Bougares, et~al.
\newblock Learning phrase representations using rnn encoder-decoder for
  statistical machine translation.
\newblock {\em arXiv preprint arXiv:1406.1078}, 2014.

\bibitem[\protect\citeauthoryear{Chung \bgroup \em et al.\egroup
  }{2015}]{chung2015gated}
Junyoung Chung, Caglar Gulcehre, Kyunghyun Cho, and Yoshua Bengio.
\newblock Gated feedback recurrent neural networks.
\newblock In {\em Proceedings of the International Conference on Machine
  Learning}, pages 2067--2075, 2015.

\bibitem[\protect\citeauthoryear{Dario \bgroup \em et al.\egroup
  }{2016}]{pmlr-v48-amodei16}
Amodei Dario, Ananthanarayanan Sundaram, Anubhai Rishita, et~al.
\newblock Deep speech 2 : End-to-end speech recognition in english and
  mandarin.
\newblock In {\em Proceedings of the International Conference on Machine
  Learning}, pages 173--182, 2016.

\bibitem[\protect\citeauthoryear{Goodfellow \bgroup \em et al.\egroup
  }{2016}]{goodfellow2016deep}
Ian Goodfellow, Yoshua Bengio, and Aaron Courville.
\newblock {\em Deep learning}.
\newblock MIT press, 2016.

\bibitem[\protect\citeauthoryear{Graves \bgroup \em et al.\egroup
  }{2014}]{graves2014neural}
Alex Graves, Greg Wayne, and Ivo Danihelka.
\newblock Neural turing machines.
\newblock {\em arXiv preprint arXiv:1410.5401}, 2014.

\bibitem[\protect\citeauthoryear{Hamilton and James}{1994}]{hamilton1994time}
Hamilton and Douglas James.
\newblock {\em Time series analysis}, volume~2.
\newblock Princeton University Press, 1994.

\bibitem[\protect\citeauthoryear{Hochreiter and
  Schmidhuber}{1997}]{hochreiter1997long}
Sepp Hochreiter and J{\"u}rgen Schmidhuber.
\newblock Long short-term memory.
\newblock {\em Neural Computation}, 9(8):1735--1780, 1997.

\bibitem[\protect\citeauthoryear{Katz}{1987}]{katz1987estimation}
Slava Katz.
\newblock Estimation of probabilities from sparse data for the language model
  component of a speech recognizer.
\newblock {\em IEEE Transactions on Acoustics, Speech, and Signal Processing},
  35(3):400--401, 1987.

\bibitem[\protect\citeauthoryear{Kim \bgroup \em et al.\egroup
  }{2017}]{ijcai2017-280}
Kyung-Min Kim, Min-Oh Heo, Seong-Ho Choi, and Byoung-Tak Zhang.
\newblock Deepstory: Video story qa by deep embedded memory networks.
\newblock In {\em Proceedings of the International Joint Conference on
  Artificial Intelligence}, pages 2016--2022, 2017.

\bibitem[\protect\citeauthoryear{Kingma and Ba}{2014}]{kingma2014adam}
Diederik~P Kingma and Jimmy Ba.
\newblock Adam: A method for stochastic optimization.
\newblock {\em arXiv preprint arXiv:1412.6980}, 2014.

\bibitem[\protect\citeauthoryear{Lang}{1995}]{lang1995newsweeder}
Ken Lang.
\newblock Newsweeder: Learning to filter netnews.
\newblock In {\em Proceedings of the International Conference on Machine
  Learning}, pages 331--339, 1995.

\bibitem[\protect\citeauthoryear{Lichman}{2013}]{Lichman:2013}
M.~Lichman.
\newblock {UCI} machine learning repository, 2013.

\bibitem[\protect\citeauthoryear{Santoro \bgroup \em et al.\egroup
  }{2016}]{santoro2016meta}
Adam Santoro, Sergey Bartunov, Matthew Botvinick, Daan Wierstra, and Timothy
  Lillicrap.
\newblock Meta-learning with memory-augmented neural networks.
\newblock In {\em International Conference on Machine Learning}, pages
  1842--1850, 2016.

\bibitem[\protect\citeauthoryear{Shazeer \bgroup \em et al.\egroup
  }{2017}]{shazeer2017outrageously}
Noam Shazeer, Azalia Mirhoseini, Krzysztof Maziarz, Andy Davis, Quoc Le,
  Geoffrey Hinton, and Jeff Dean.
\newblock Outrageously large neural networks: The sparsely-gated
  mixture-of-experts layer.
\newblock {\em arXiv preprint arXiv:1701.06538}, 2017.

\bibitem[\protect\citeauthoryear{Sutskever \bgroup \em et al.\egroup
  }{2014}]{sutskever2014sequence}
Ilya Sutskever, Oriol Vinyals, and Quoc~V Le.
\newblock Sequence to sequence learning with neural networks.
\newblock In {\em Advances in Neural Information Processing Systems}, pages
  3104--3112, 2014.

\bibitem[\protect\citeauthoryear{Vaswani \bgroup \em et al.\egroup
  }{2017}]{NIPS2017_7181}
Ashish Vaswani, Noam Shazeer, Niki Parmar, Jakob Uszkoreit, Llion Jones,
  Aidan~N Gomez, et~al.
\newblock Attention is all you need.
\newblock In {\em Proceedings of the Advances in Neural Information Processing
  Systems}, pages 6000--6010. 2017.

\bibitem[\protect\citeauthoryear{Weston \bgroup \em et al.\egroup
  }{2014}]{weston2014memory}
Jason Weston, Sumit Chopra, and Antoine Bordes.
\newblock Memory networks.
\newblock {\em arXiv preprint arXiv:1410.3916}, 2014.

\bibitem[\protect\citeauthoryear{Williams and
  Hinton}{1986}]{williams1986learning}
DRGHR Williams and Geoffrey Hinton.
\newblock Learning representations by back-propagating errors.
\newblock {\em Nature}, 323(6088):533--538, 1986.

\bibitem[\protect\citeauthoryear{Yi \bgroup \em et al.\egroup
  }{2018}]{ijcai2018-633}
Xiaoyuan Yi, Maosong Sun, Ruoyu Li, and Zonghan Yang.
\newblock Chinese poetry generation with a working memory model.
\newblock In {\em Proceedings of the International Joint Conference on
  Artificial Intelligence}, pages 4553--4559, 2018.

\bibitem[\protect\citeauthoryear{Zaremba \bgroup \em et al.\egroup
  }{2014}]{zaremba2014recurrent}
Wojciech Zaremba, Ilya Sutskever, and Oriol Vinyals.
\newblock Recurrent neural network regularization.
\newblock {\em arXiv preprint arXiv:1409.2329}, 2014.

\bibitem[\protect\citeauthoryear{Zilly \bgroup \em et al.\egroup
  }{2016}]{zilly2016recurrent}
Julian~Georg Zilly, Rupesh~Kumar Srivastava, Jan Koutn{\'\i}k, and J{\"u}rgen
  Schmidhuber.
\newblock Recurrent highway networks.
\newblock {\em arXiv preprint arXiv:1607.03474}, 2016.

\end{thebibliography}

\end{document}